\documentclass[11pt]{article}
\usepackage[utf8]{inputenc}
\usepackage[T1]{fontenc}
\usepackage{mathptmx}
\usepackage[letterpaper,margin=1in]{geometry}
\usepackage{graphicx}
\usepackage{array}
\usepackage[font=bf,labelsep=period]{caption}
\begin{document}
\begin{center}
{\large\textbf{Toward Open Room-Scale Scanning with Consumer LiDAR\\[0.2em]
via Depth--Inertial Odometry}}\\[1.5em]
Minsuk Jang\\[0.8em]
{\small KAIST, Daejeon, Republic of Korea. Email: minsukjang@kaist.ac.kr}
\end{center}

\begin{abstract}
\noindent Room scanning with consumer LiDAR today depends on ARKit, whose tracking and map update are a black box: an application receives depth and poses but cannot inspect how they were produced, so an error in a scan cannot be traced to its cause. This paper examines whether an open alternative, one whose every stage can be inspected, is feasible on the device. Mobile-LIO is a depth--inertial odometry pipeline that uses the phone's depth and inertial data and no ARKit poses. On seven ScanNet++ iPhone recordings it completed every scene with 0.2--2.0 m deviation from the ARKit-derived reference, where six open LiDAR--inertial estimators run as configured for dedicated scanners had at least 4.46 m in 41 of 42 runs (two of the six also failed a noise-free adapter check, and Mobile-LIO has a speed cap they lack; the comparison is not controlled and does not rank algorithms). On an iPhone 15 Pro Max, a build with corrected phone code kept up with the sensor in 14 of 15 replay runs of three 7--13 s recordings; the dataset results use the estimator before those corrections. Laser-scanned meshes show how far this is from the closed pipeline. Within 3 m, recorded depth differed from the mesh by a median of 1--5 cm; on three held-out scenes, maps fused with ARKit poses that the dataset scales and aligns to the mesh lay a median 1.8--2.3 cm from it, and Mobile-LIO maps 8--17 cm. Because the pipeline is open, the gap could be examined: in the scene with the largest error, noise-free rendered depth cut the deviation from 1.97 m to 0.04 m, so recorded depth, whose error reached 20--54 cm at 5--7 m, contributed alongside the estimator. Open room-scale scanning on the device is therefore feasible but not yet as accurate as ARKit-pose maps, which themselves indicatively reach only the coarsest USIBD accuracy level (LOA10).
\end{abstract}

\section{Introduction}

\noindent Geometric digital twins of buildings lose value unless their geometry is reacquired as the asset changes (Osadcha et al. 2023; Davletshina et al. 2026). Terrestrial laser scanning delivers the accuracy that scan-to-BIM needs, but each update is costly. In one comparison over about 350 m\textsuperscript{2} of indoor space, a terrestrial scanning and manual modeling workflow took 20 h and a tablet-based workflow 32 min, with larger geometric discrepancies (Ersoz and Bosch\'e 2026). Low-cost, repeatable capture is therefore attractive.

Studies in geomatics, geoscience, and civil engineering have evaluated the output of LiDAR-equipped iPhones and iPads for indoor mapping and building documentation (Teo and Yang 2023; Askar and Sternberg 2023; Abdel-Majeed et al. 2024; Treccani et al. 2024) and for geoscience surveys (Luetzenburg et al. 2021). The pose tracking and map update behind these outputs belong to the proprietary ARKit framework, which exposes depth, confidence, and camera pose but not its internal decisions. When an export shows depth smeared across object edges or a corridor loop that does not close (Jang and Yu 2026), the user cannot trace the error to a stage.

An application-level accuracy number therefore mixes the error of the depth sensor, the error of the estimated pose, and the choices of the fusion software. Open-source LiDAR--inertial odometry (LIO), which estimates the pose of a moving device from range and motion sensors, offers inspectable estimation, yet representative systems are designed for dedicated scanners (Xu et al. 2022; Chen et al. 2023). Smartphone depth is a different observation regime: a 256 $\times$ 192 depth image with a three-level confidence map and no per-point timing.

This paper examines whether room-scale scanning is feasible with an open pipeline, meaning one whose every stage is specified and can be inspected, and measures how far such a pipeline is from the closed one by separating those terms. It makes five contributions: (1) a measurement of recorded iPhone depth error against a laser-scanned mesh as a function of range; (2) an error budget on the same recordings, giving the surface error when the same depth is fused with mesh-aligned ARKit poses and with the poses of Mobile-LIO, an inspectable depth--inertial pipeline whose settings are listed in the Appendix, and the deviation of Mobile-LIO from that ARKit-derived reference; (3) a controlled experiment, made possible by the inspectable pipeline, that fixes the estimator and varies only the depth input; (4) a comparison with six open LiDAR--inertial estimators on the same input; and (5) supporting checks: an on-device replay before and after code corrections and qualitative scans exported by the phone application. The results are related to the accuracy levels used to specify building documentation. The scope is single indoor scenes and local estimation without loop closure (the correction of drift when a route returns to a known place); on-device evidence comes from three short recordings of one room replayed on one phone (timing and agreement with ARKit over a few seconds), qualitative scans exported by the phone application, and one corridor compared with a scanner map; it is separate from the held-out evaluation.

\section{Related Work}

\medskip\noindent\textbf{Smartphone LiDAR in the built environment.} Teo and Yang (2023) and Abdel-Majeed et al. (2024) evaluated Apple devices for indoor mapping, Askar and Sternberg (2023) compared scanning applications for building documentation, and Luetzenburg et al. (2021) characterized the iPhone 12 Pro sensor for geoscience surveys. Treccani et al. (2024) tested five iOS applications against a terrestrial scanner in a corridor and a street and found errors from about 1 cm to about 10 cm. For as-built BIM updating, Yang et al. (2024) replaced terrestrial scanning with a low-cost mobile laser scanner and an inertial unit. D\'{\i}az-Vilari\~no et al. (2022) assessed precision, correctness, and coverage of iPad Pro scans of two rooms, and Teppati Los\`e et al. (2022) and Abdelghany et al. (2026) tested the sensor for heritage documentation. The common protocol registers the exported cloud to a terrestrial scan, usually by iterative closest point (ICP), and reports cloud-to-cloud distance and sometimes dimensions. Such comparisons report end-product accuracy and cannot separate depth error from tracking error; the present study separates depth error from the tracking error of an open depth--inertial estimator, though not from ARKit's own.

\medskip\noindent\textbf{LiDAR--inertial odometry and construction SLAM.} FAST-LIO2 couples inertial propagation with direct point-to-map registration in an iterated Kalman filter (Xu et al. 2022); Faster-LIO, Point-LIO, and DLIO are further open estimators (Bai et al. 2022; He et al. 2023; Chen et al. 2023). In construction, the ConSLAM dataset benchmarks handheld SLAM (Trzeciak et al. 2023) and Keitaanniemi et al. (2023) analyzed drift in indoor SLAM laser scanning; both use dedicated scanners. On a phone, Jin and Ye (2023) demonstrated visual--LiDAR--inertial odometry on an iPhone 12 Pro. Jang and Yu (2026) introduced Mobile-LIO, with confidence-aware selection and a budgeted filter for iPhone depth together with velocity-feedback and loop-correction components. The present paper evaluates only its local estimator, without those components, and adds the settings listed in the Appendix, a held-out evaluation against a laser reference, and the error budget.

\section{Method}

\noindent Figure 1 shows the pipeline. One C++ estimator serves an iPhone application and a desktop replay tool that re-runs recorded sequences; neither uses ARKit camera poses. The estimator predicts the phone's motion from its motion sensors and corrects the prediction by how well each new depth image lines up with surfaces already mapped.

\begin{figure}[htbp]
\centering
\includegraphics[width=5.60in]{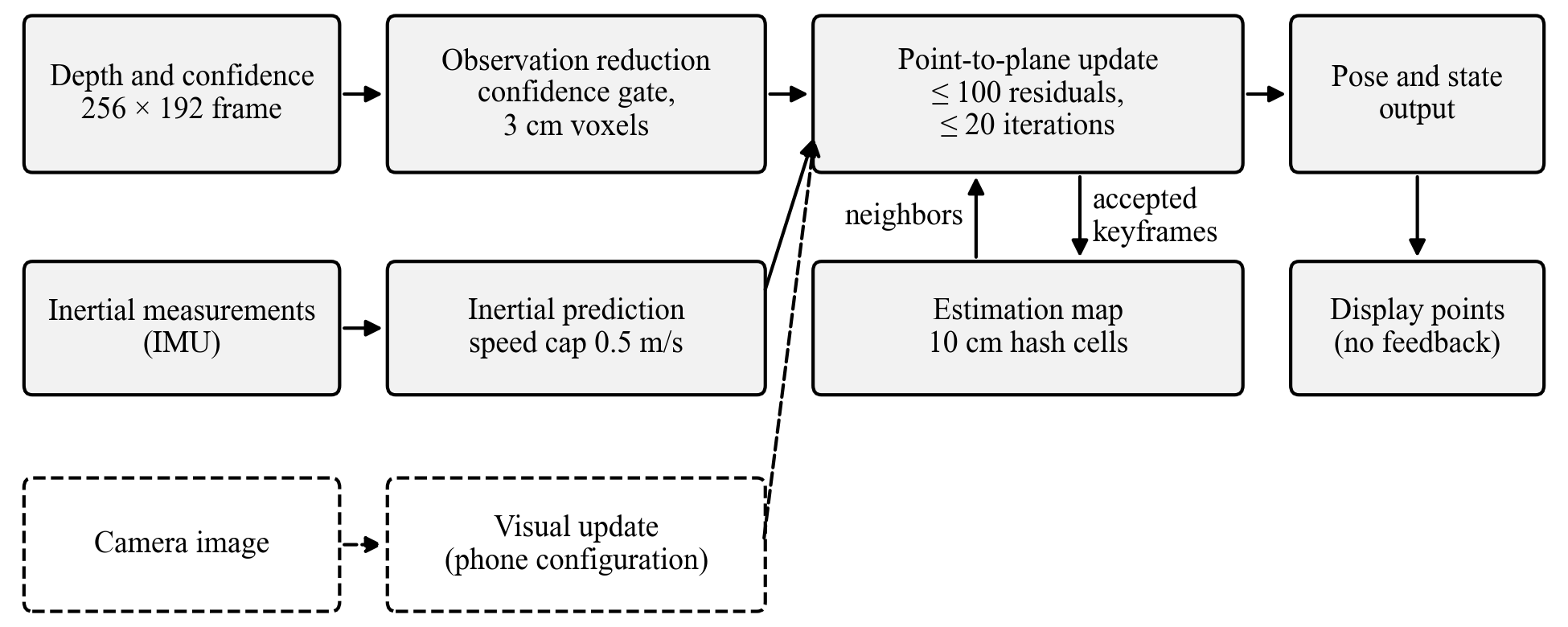}
\caption{Mobile-LIO pipeline. Solid boxes form the depth--inertial configuration; dashed boxes are added in the phone configuration.}
\end{figure}

\medskip\noindent\textbf{Input reduction.} Each 256 $\times$ 192 depth frame is back-projected with a stride of four pixels, which gives at most 3,072 points. Samples with confidence level 0, nonfinite coordinates, or range beyond a limit (10 m or 3 m, depending on the configuration) are discarded. The surviving points are grouped into 3 cm voxels, and each voxel is replaced by its centroid and mean confidence.

\medskip\noindent\textbf{Pose estimation.} Motion-sensor samples propagate orientation, position, velocity, sensor biases, and gravity between depth frames. Propagated speed is clipped at 0.5 m/s; because depth does not observe velocity directly, the cap keeps the velocity estimate bounded, but it can also clip faster true motion. Each reduced point is then compared with a plane fitted to the mapped points within a 20 cm radius (at least six, drawn from a map that keeps up to 20 points per 10 cm cell), and its distance to that plane is the residual. Residuals above 10 cm are rejected, the rest are weighted by confidence with a noise level of 2 cm, and at most 100 enter a Kalman update that is repeated up to 20 times with fresh plane matches; the gain uses the predicted covariance throughout, and the covariance is updated once, after the last iteration. Unlike the iterated filter in FAST-LIO2, the repeated update does not pull each iterate back toward the predicted state, so it is a repeated Kalman correction and not an iterated filter. Every result in the Results section uses this update; the Appendix lists the remaining settings. A total correction above 1 m or 0.5 rad is discarded.

\medskip\noindent\textbf{Maps and recovery.} The first five frames seed the estimation map at the propagated pose. After seeding, points enter the map only on a keyframe (5 cm or 2$^\circ$ of motion) whose update was accepted. Ten consecutive failed updates clear and reseed the map; displayed points never feed back into estimation.

\medskip\noindent\textbf{Two configurations.} The depth--inertial configuration uses depth to 10 m and motion-sensor (inertial measurement unit, IMU) data only; its settings were frozen before the held-out scenes were run. The phone configuration follows the inputs found in the application source: a 3 m range limit and a visual update from tracked image features (it was also run without the visual update), here with the initial gyroscope-bias estimate disabled because the recordings do not start at rest. It was run on the held-out scenes after the first results were known and is therefore a post hoc second look. Its replay maps the inertial axes into the ARKit camera frame of the points; the application source as it stood before this study (as received) has them rotated 180$^\circ$ about the optical axis, which a later check on the phone confirmed.

\section{Experimental Design}

\medskip\noindent\textbf{Data.} ScanNet++ provides iPhone depth, images, inertial data, ARKit poses, a laser-scanned mesh, and the same ARKit poses scaled and transformed into the mesh frame for each scene (Yeshwanth et al. 2023); the latter are called the dataset poses here and are the reference for all trajectory metrics. They are not an independent ground truth. Of 50 locally available scenes, seven had been used in development or earlier evaluation and were excluded, so the evaluated scenes are held out. A rule recorded before any run selected the first three remaining scene identifiers that passed predeclared checks on file integrity, duration, frame gaps, inertial consistency, and initial motion. No scene was replaced after its result was seen. Every sixth frame was used, giving 10 Hz depth with 60 Hz inertial data. Scenes S1 to S3 (ScanNet++ scenes 286b55a2bf, 31a2c91c43, and 578511c8a9) have 900, 1,270, and 2,549 frames, path lengths of 22, 20, and 92 m, and 27, 34, and 347 m\textsuperscript{2} of observed mesh surface. The converted data carry no native confidence, so every sample received level 2.

\medskip\noindent\textbf{Depth error.} For the three held-out and seven development scenes, depth was rendered from the laser mesh at the dataset poses and compared pixel by pixel with the recorded depth on 300 to 1,446 frames per scene. Recorded depth is the depth map delivered by ARKit, not raw sensor output. Errors were binned by recorded range.

\medskip\noindent\textbf{Pose sources and trajectory metrics.} The pose sources were the dataset poses, the raw ARKit poses stored with each frame (a consistency check, since the dataset poses derive from them), the two Mobile-LIO configurations, and FAST-LIO2, Faster-LIO, Point-LIO, and DLIO given the same depth points, inertial samples, and interval through an adapter with identity extrinsics and zero per-point time offsets. On noise-free synthetic sequences the adapter gave errors below 1 cm for all four. Two further open estimators, iG-LIO and LIO-Livox, showed large errors on some of these checks; they are reported in Table 6 with that caveat. Absolute trajectory error (ATE) is the translational root mean square error after one rigid alignment without scale. A run is said to lose tracking when its translation error over 1 s first exceeds 0.5 m. A run that loses tracking and ends with ATE above 4 m is called diverged.

\medskip\noindent\textbf{Surface metrics.} For each scene the same depth frames were fused with dataset poses and with Mobile-LIO poses, using a 0.25--4.0 m range, every second depth pixel, 2 cm voxel centroids, and at least two hits per voxel. The Mobile-LIO trajectory was placed in the dataset frame by the rigid alignment used for ATE, with no map-to-mesh registration, so the distances include residual global misalignment as well as drift. Exact point-to-triangle distance to the laser mesh was computed for every map point. Coverage is the share of mesh samples observed in the dataset-pose map that lie within 5 cm of the evaluated map.

\medskip\noindent\textbf{Controlled depth input.} On S3, the depth--inertial estimator was held fixed while only its depth input changed: recorded depth limited to 10, 5, or 4 m; recorded depth keeping only pixels within 5 cm of the mesh, a diagnostic that needs the reference; and noise-free depth rendered from the mesh at the phone's field of view and at two wider fields of view.

\section{Results}

\medskip\noindent\textbf{Recorded depth is accurate within 3 m.} Recorded depth was accurate at close range and unreliable far away (Figure 2, Table 1). Across the ten scenes the median absolute error per scene was 0.9--2.2 cm at 0--1 m, 1.6--4.8 cm at 2--3 m, 2.9--10.3 cm at 4--5 m, and 19.7--53.5 cm at 5--7 m. Within 3 m, where about 90\% of the 14.2 million compared pixels lay, recorded depth read short by a median of up to 2 cm, and the 90th percentile of the error was 3--32 cm, so even close-range depth has a tail. S1 and S2 contained almost no pixels beyond 3 m, whereas 16.6\% of the pixels in S3 lay beyond 4 m.

\begin{figure}[htbp]
\centering
\includegraphics[width=6.50in]{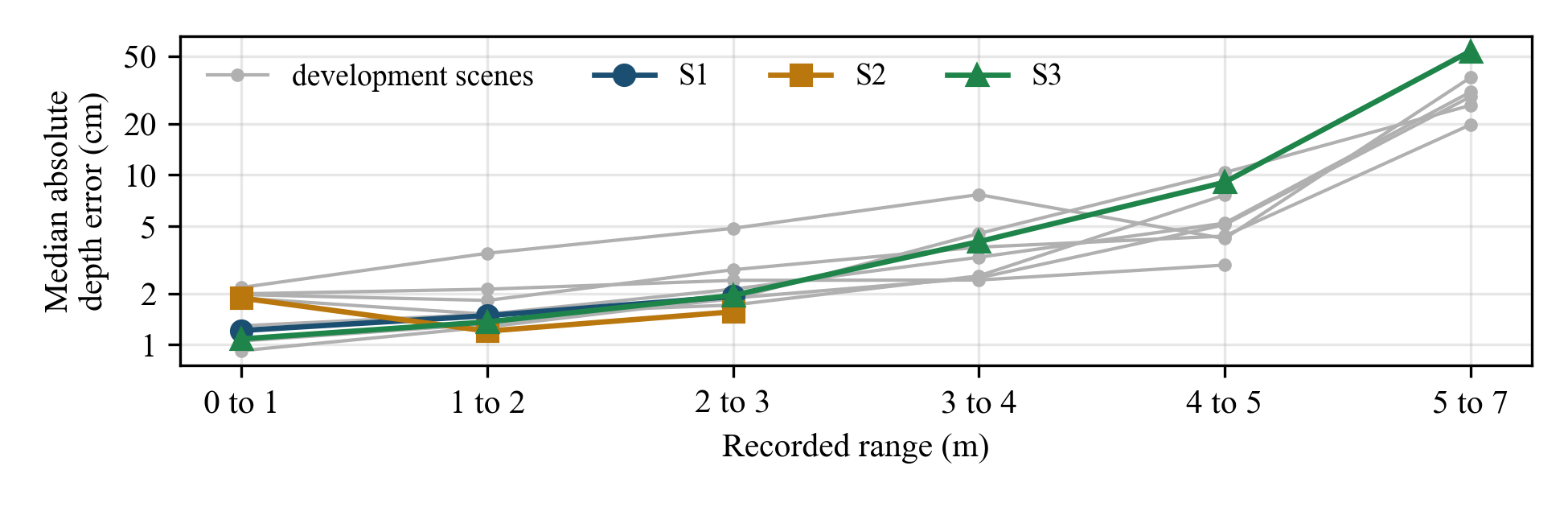}
\caption{Median absolute error of recorded iPhone depth against depth rendered from the laser mesh, by range, for ten scenes.}
\end{figure}

\begin{table}[htbp]
\centering
\caption{Error of recorded depth by range over ten scenes. Error columns give the range of per-scene values; the signed median is recorded minus rendered depth.}
\small
\begin{tabular}{|>{\centering\arraybackslash}p{0.91in}|>{\centering\arraybackslash}p{0.91in}|>{\centering\arraybackslash}p{0.91in}|>{\centering\arraybackslash}p{0.91in}|>{\centering\arraybackslash}p{0.91in}|>{\centering\arraybackslash}p{0.91in}|}
\hline
\textbf{Range (m)} & \textbf{Pixels (\%)} & \textbf{Signed median (cm)} & \textbf{Absolute median (cm)} & \textbf{P90 (cm)} & \textbf{Above 5 cm (\%)} \\ \hline
0--1 & 23.9 & $-$1.7 to 0.0 & 0.9--2.2 & 2.8--26.5 & 3--32 \\ \hline
1--2 & 48.6 & $-$1.6 to $-$0.5 & 1.2--3.5 & 3.7--28.3 & 5--41 \\ \hline
2--3 & 17.2 & $-$2.0 to $-$0.6 & 1.6--4.8 & 4.0--32.3 & 6--49 \\ \hline
3--4 & 5.7 & $-$1.8 to $-$0.2 & 2.4--7.6 & 8.3--42.2 & 21--58 \\ \hline
4--5 & 2.0 & $-$3.8 to $-$1.1 & 2.9--10.3 & 10.3--122.0 & 30--68 \\ \hline
5--7 & 2.4 & $-$45.1 to 17.0 & 19.7--53.5 & 50.9--160.3 & 82--94 \\ \hline
\end{tabular}
\end{table}

\medskip\noindent\textbf{Mobile-LIO deviated from the ARKit-derived reference by 0.2 to 2.0 m.} Raw ARKit poses differed from the dataset poses by 3--8 cm after rigid alignment (Table 2); because the dataset poses are those ARKit poses with a scale and transform applied, this measures that adjustment and not ARKit's accuracy. The depth--inertial Mobile-LIO configuration completed all three recordings with ATE of 0.409, 0.200, and 1.973 m. The phone configuration with visual updates reached 0.435, 0.170, and 0.590 m; without visual updates it reached 2.920 m in S3, so the improvement in the large scene points to the visual update, not the range limit (Figure 3). By the tracking-loss criterion, the depth--inertial configuration first exceeded the threshold at 19 s in S1 and 35 s in S3 and never in S2; it did not diverge, but 13\% (S1) and 19\% (S3) of its 1 s intervals exceeded 0.5 m. The four estimators that passed the adapter check diverged (all six are compared in Table 6): DLIO exceeded it within 5 s, FAST-LIO2 after 23--73 s, Faster-LIO after 35--81 s, and Point-LIO after 32--67 s, and all ended with ATE between 7 m and tens of kilometers. That comparison is not controlled: Mobile-LIO has a 0.5 m/s speed cap that the others lack, no recording starts at rest as DLIO's calibration assumes, and on development scenes removing the cap made three of seven Mobile-LIO runs diverge.

\begin{table}[htbp]
\centering
\caption{Deviation from the dataset poses (ATE, m) on the held-out scenes. The dataset poses are derived from the raw ARKit poses.}
\small
\begin{tabular}{|>{\centering\arraybackslash}p{1.44in}|>{\centering\arraybackslash}p{1.44in}|>{\centering\arraybackslash}p{1.44in}|>{\centering\arraybackslash}p{1.44in}|}
\hline
\textbf{Pose source} & \textbf{S1} & \textbf{S2} & \textbf{S3} \\ \hline
ARKit, raw (same provenance as the reference) & 0.030 & 0.033 & 0.082 \\ \hline
Mobile-LIO, depth--inertial & 0.409 & 0.200 & 1.973 \\ \hline
Mobile-LIO, phone configuration, no visual update & 0.441 & 0.230 & 2.920 \\ \hline
Mobile-LIO, phone configuration & 0.435 & 0.170 & 0.590 \\ \hline
\end{tabular}
\end{table}

\begin{figure}[htbp]
\centering
\includegraphics[width=6.50in]{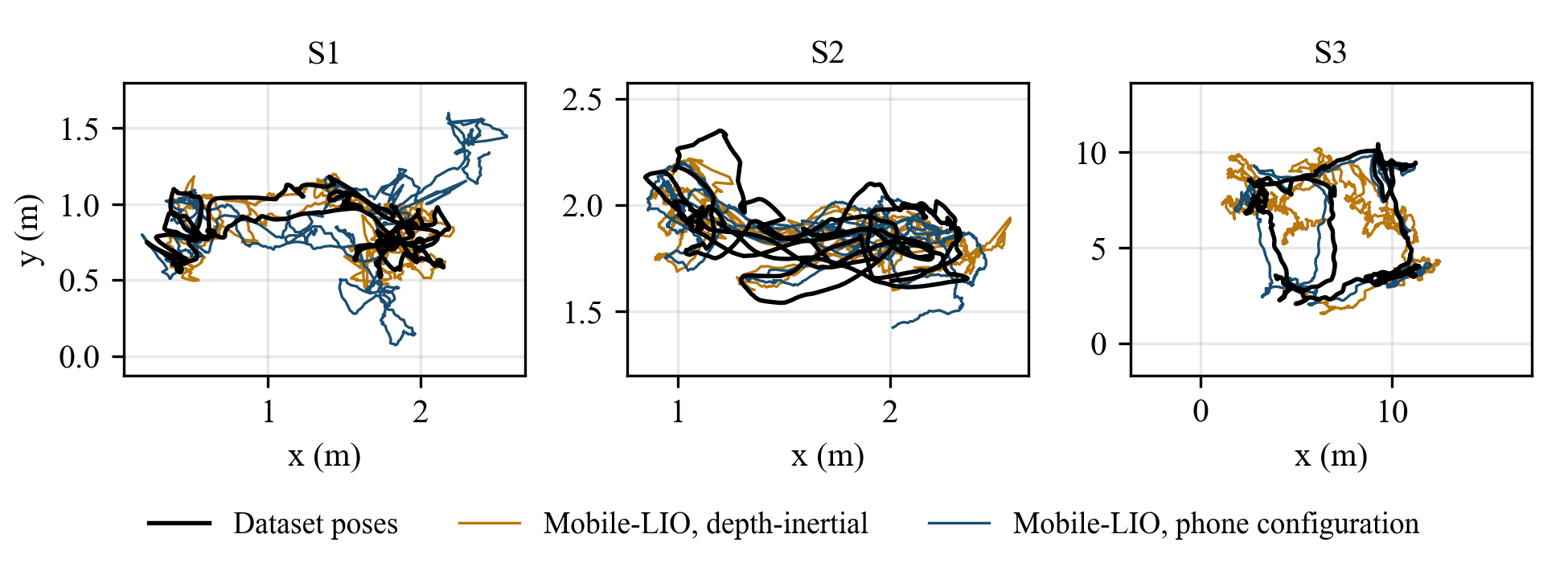}
\caption{Top-down trajectories on the held-out scenes after rigid alignment to the dataset poses.}
\end{figure}

\medskip\noindent\textbf{ARKit-pose maps lay about 2 cm from the laser mesh; Mobile-LIO maps 8 to 17 cm.} With Mobile-LIO poses, pose error, not depth, set most of the median surface error (Table 3, Figure 4). With dataset poses, which are ARKit's poses in the mesh frame, the median distance was 1.84--2.29 cm; this is what ARKit tracking with the dataset's scale correction achieves with this depth and fusion rule. The USIBD Level of Accuracy bands are specified at the 95\% confidence level: LOA20 spans 15 mm to 5 cm and LOA10 is coarser than 5 cm (USIBD 2016). In the dataset-pose maps 17--27\% of points lay beyond 5 cm, so the maps fused with aligned ARKit poses indicatively correspond to LOA10; with Mobile-LIO poses 68--83\% lay beyond 5 cm. Nearest-surface distance over all map points is not a formal LOA validation, so this placement is indicative. With depth--inertial Mobile-LIO poses it rose to 8.35--17.48 cm and coverage at 5 cm fell to 34--58\%. The phone configuration improved S3 (median 17.48 to 12.52 cm, 90th percentile 63.99 to 34.89 cm) and coverage in S2, but it was worse in S1, so it cannot be called better overall. The S2 dataset-pose map has a 90th percentile of 42 cm; Figure 4 shows points through an opening where the mesh may not extend, so that tail may reflect mesh coverage and not depth error. It was not masked.

\begin{table}[htbp]
\centering
\caption{Distance of fused maps to the laser-scanned mesh (P90: 90th percentile) for dataset poses (mesh-aligned ARKit) and the two Mobile-LIO configurations. Coverage is 100\% by definition for dataset poses.}
\small
\begin{tabular}{|>{\centering\arraybackslash}p{0.91in}|>{\centering\arraybackslash}p{0.91in}|>{\centering\arraybackslash}p{0.91in}|>{\centering\arraybackslash}p{0.91in}|>{\centering\arraybackslash}p{0.91in}|>{\centering\arraybackslash}p{0.91in}|}
\hline
\textbf{Scene} & \textbf{Pose source} & \textbf{Median (cm)} & \textbf{P90 (cm)} & \textbf{Points within 5 cm (\%)} & \textbf{Coverage at 5 cm (\%)} \\ \hline
S1 & Dataset & 1.84 & 8.44 & 82.8 & 100.0 \\ \hline
S1 & Depth--inertial & 8.35 & 25.49 & 31.7 & 57.7 \\ \hline
S1 & Phone configuration & 13.75 & 52.90 & 21.3 & 59.6 \\ \hline
S2 & Dataset & 2.29 & 42.44 & 73.1 & 100.0 \\ \hline
S2 & Depth--inertial & 9.28 & 36.95 & 25.6 & 51.0 \\ \hline
S2 & Phone configuration & 9.32 & 27.65 & 28.0 & 62.6 \\ \hline
S3 & Dataset & 2.15 & 8.55 & 78.3 & 100.0 \\ \hline
S3 & Depth--inertial & 17.48 & 63.99 & 16.9 & 33.7 \\ \hline
S3 & Phone configuration & 12.52 & 34.89 & 21.2 & 38.6 \\ \hline
\end{tabular}
\end{table}

\begin{figure}[htbp]
\centering
\includegraphics[width=5.60in]{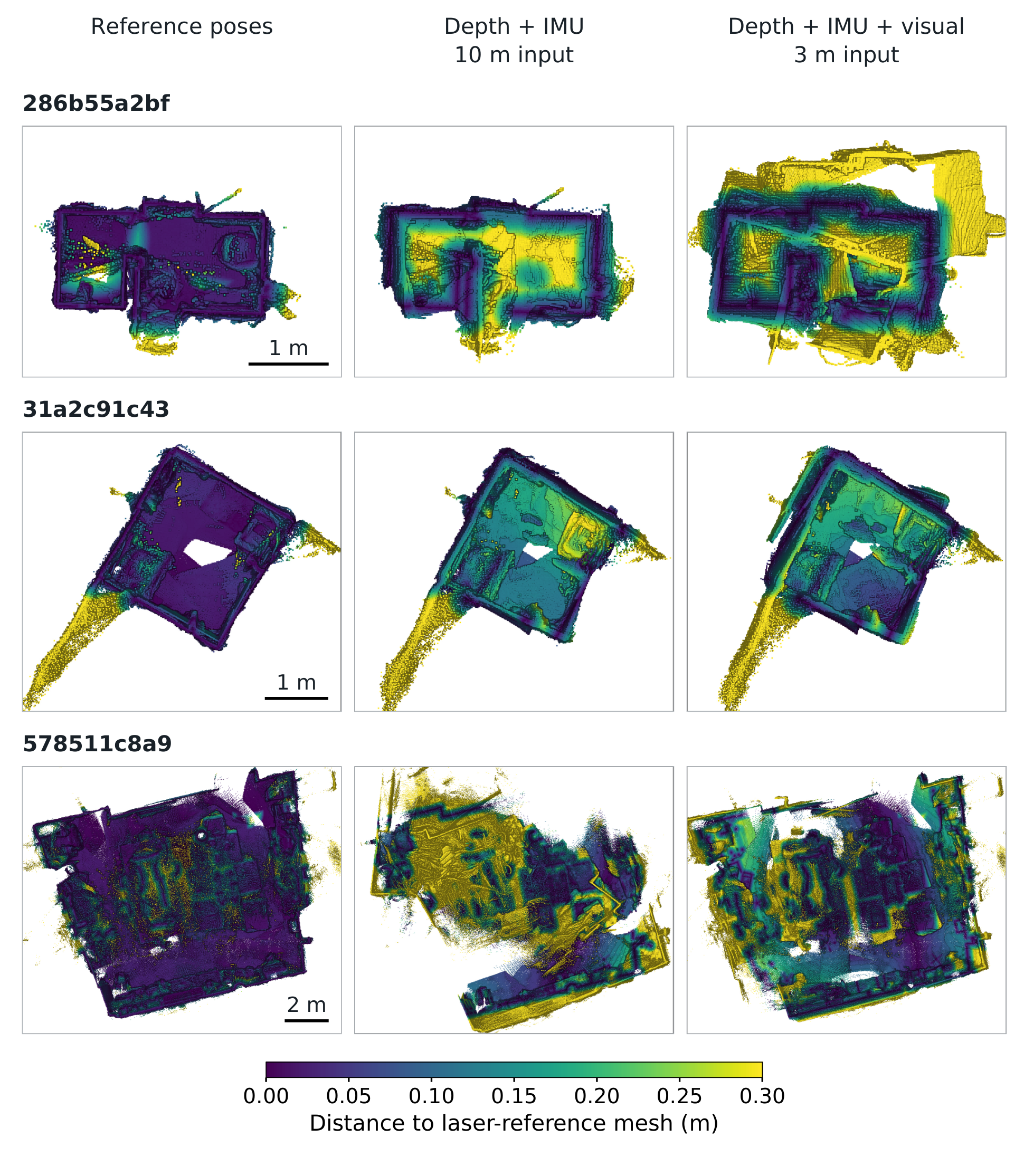}
\caption{Distance of fused maps to the laser mesh for S1 to S3 (top to bottom) with dataset poses (mesh-aligned ARKit; labeled reference poses), depth--inertial Mobile-LIO poses, and phone-configuration Mobile-LIO poses (column heads: depth and IMU with 10 m input; depth, IMU, and visual with 3 m input). Row labels are the ScanNet++ scene identifiers of S1 to S3. Top-down views; the color scale saturates at 0.3 m. Data: Yeshwanth et al. (2023).}
\end{figure}

\medskip\noindent\textbf{Idealized depth removed the large error in S3; filtering recorded depth did not.} With the estimator fixed, limiting recorded depth to 5 or 4 m did not help (Figure 5), so range alone does not explain the large error. Keeping only pixels within 5 cm of the mesh cut ATE from 1.973 to 0.594 m, and noise-free rendered depth at the phone's field of view cut it to 0.042 m. With the two wider fields of view the values were 0.026 and 0.016 m. Rendered depth removes edge, dropout, and reflective-surface errors together with range-dependent error, and it is dense and exactly consistent with the dataset poses, so it is an idealized bound and the experiment does not say which of these matters most; it covers one scene. In the per-frame logs, position error measured from the start pose stayed at 0.13--0.20 m for the first 60 s (the 1 s criterion was first exceeded at 35 s) and grew when the share of far pixels jumped to 30\%, which is consistent with a role for far-range depth but does not isolate it.

\begin{figure}[htbp]
\centering
\includegraphics[width=6.50in]{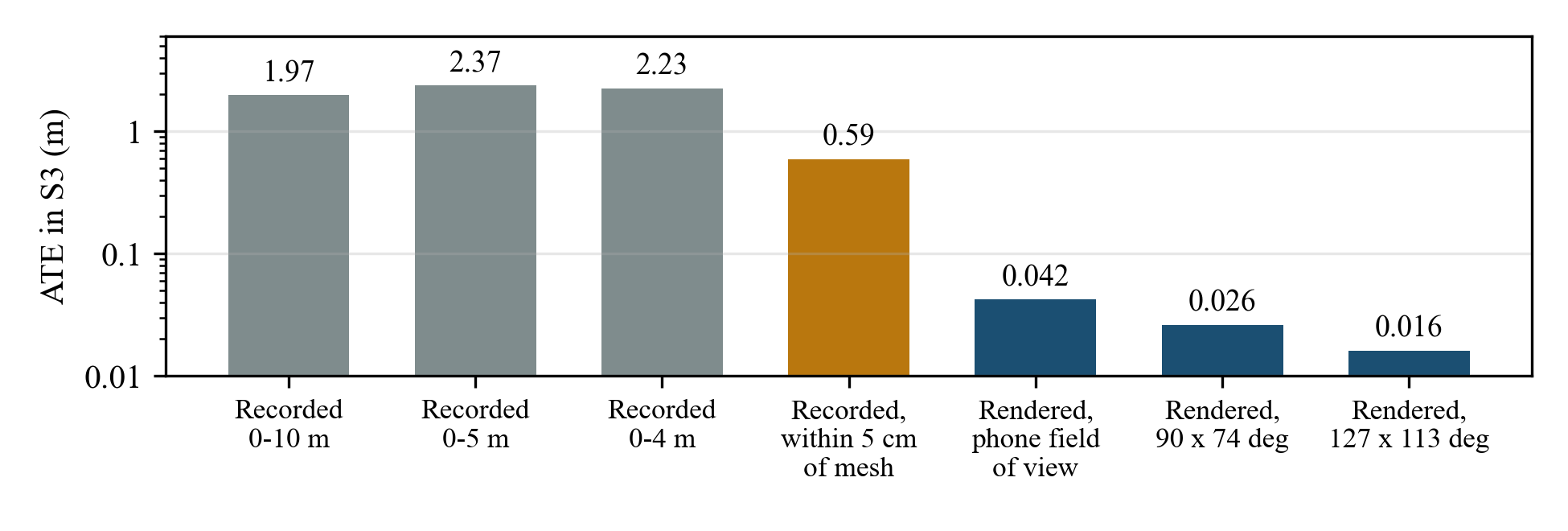}
\caption{Trajectory error in S3 with the estimator fixed and only the depth input changed. Rendered depth is noise-free depth from the laser mesh.}
\end{figure}

\medskip\noindent\textbf{Processing cost.} For the depth--inertial configuration on a laptop with a Ryzen 5 5600H processor, correspondence search took 95.0--99.6\% of mean frame time across residual budgets of 50 to 200, and median frame time at the default budget was 188--489 ms, slower than the 10 Hz input. Timing on the phone is reported in the next section.

\section{On-Device Replay}

\noindent The results above come from desktop replay of dataset recordings with the estimator of the Method section. This section reports what that estimator does on a phone and what changed when its phone code was corrected. It is a post hoc check on short recordings, separate from the held-out evaluation, and the corrected build described here was not run on the dataset.

\medskip\noindent\textbf{Recordings and protocol.} Three hand-held scans of one room were recorded on an iPhone 15 Pro Max (6.9, 8.4, and 12.7 s; 0.4--0.9 m of device motion; 3 m range limit; depth delivered at 60 Hz in two and 30 Hz in one). A recording stores exactly what the application passes to the estimator: 3,072 depth points per frame, the 480 $\times$ 360 gray image and depth map used by the visual update, and motion-sensor samples. For timing, the phone fed each recording back to a fresh estimator at the recorded pace, five times per recording and build, in a release build with no display load; frames that arrive while the estimator is busy are dropped, as in a live scan. Delay is measured from delivery of a frame to its pose, plus the 33 ms median capture-to-delivery delay measured in live sessions. A rule fixed before the runs calls a run able to keep up with the sensor if at least 90\% of delivered frames are processed and the 95th-percentile delay is at most 100 ms. Positions are compared with the ARKit positions logged with each recording after one rigid alignment. ARKit uses the same camera and motion sensors and is not ground truth, so agreement with it over less than 1 m of motion shows consistency, not accuracy; the estimator never reads ARKit poses.

\medskip\noindent\textbf{Two builds.} The as-received build is the application source as it stood before this study. The corrected build differs in three ways. First, motion-sensor samples are mapped from the device frame (x right, y up, z toward the user in portrait) into the ARKit camera frame of the depth points as ($-$y, x, z); the as-received build used (y, $-$x, z), which is rotated 180$^\circ$ about the optical axis. In four live sessions with the as-received build, the estimated up direction differed from ARKit's by 165--177$^\circ$ at the start; in one session with the corrected mapping it differed by 0.2$^\circ$. Second, the repeated correction includes the prior term of an iterated Kalman filter, which pulls each iterate back toward the predicted state. Third, plane fits are attempted for about 300 evenly spaced candidate points per iteration, not all reduced points, and iteration stops when a step falls below 0.2 mm and 0.2 mrad; these two were changed together and are not separated.

\medskip\noindent\textbf{The as-received build could not keep up; the corrected build could.} The as-received build took a median 192 ms per keyframe, processed 24\% of delivered frames, and no run met the rule (Table 4, Figure 6). Its median root-mean-square position difference from ARKit was 23.5 cm on recordings in which the device moved less than 1 m. The corrected build took a median 18 ms per keyframe, processed 98\% of frames with a 95th-percentile delay of 65 ms, and 14 of 15 runs met the rule; the remaining run processed 89.9\% of frames. Registration runs only on keyframes, 13--35\% of delivered frames depending on the recording, and the other frames take less time, which is why a median keyframe time slightly above the 16.7 ms frame interval still allows nearly all frames to be processed. The corrected build's median position difference from ARKit was 1.8 cm (0.8--4.5 cm across runs). In two live scans of about 7 s with the same estimator settings and with display load, 83\% and 95\% of frames were processed, so the replay is somewhat optimistic about a live scan.

\begin{table}[htbp]
\centering
\caption{On-device replay on an iPhone 15 Pro Max: median (range) of per-run values over 15 runs (three recordings, five runs each).}
\small
\begin{tabular}{|>{\centering\arraybackslash}p{1.96in}|>{\centering\arraybackslash}p{1.96in}|>{\centering\arraybackslash}p{1.96in}|}
\hline
\textbf{Measure} & \textbf{As-received build} & \textbf{Corrected build} \\ \hline
Frames processed (\%) & 24.4 (18.8--36.9) & 97.5 (89.9--100.0) \\ \hline
Time per keyframe, median (ms) & 192 (131--215) & 18.0 (15.4--20.1) \\ \hline
Time per keyframe, 95th percentile (ms) & 303 (264--342) & 36 (32--55) \\ \hline
Registration time, median (ms) & 135 (87--151) & 4.0 (3.5--4.8) \\ \hline
Delay to pose, 95th percentile (ms) & 301 (246--399) & 65 (55--72) \\ \hline
Position difference from ARKit, RMS (cm) & 23.5 (14.1--45.0) & 1.8 (0.8--4.5) \\ \hline
Estimated path / ARKit path & 5.75 (3.77--13.26) & 1.23 (1.15--1.32) \\ \hline
Runs meeting the rule & 0 of 15 & 14 of 15 \\ \hline
\end{tabular}
\end{table}

\begin{figure}[htbp]
\centering
\includegraphics[width=6.50in]{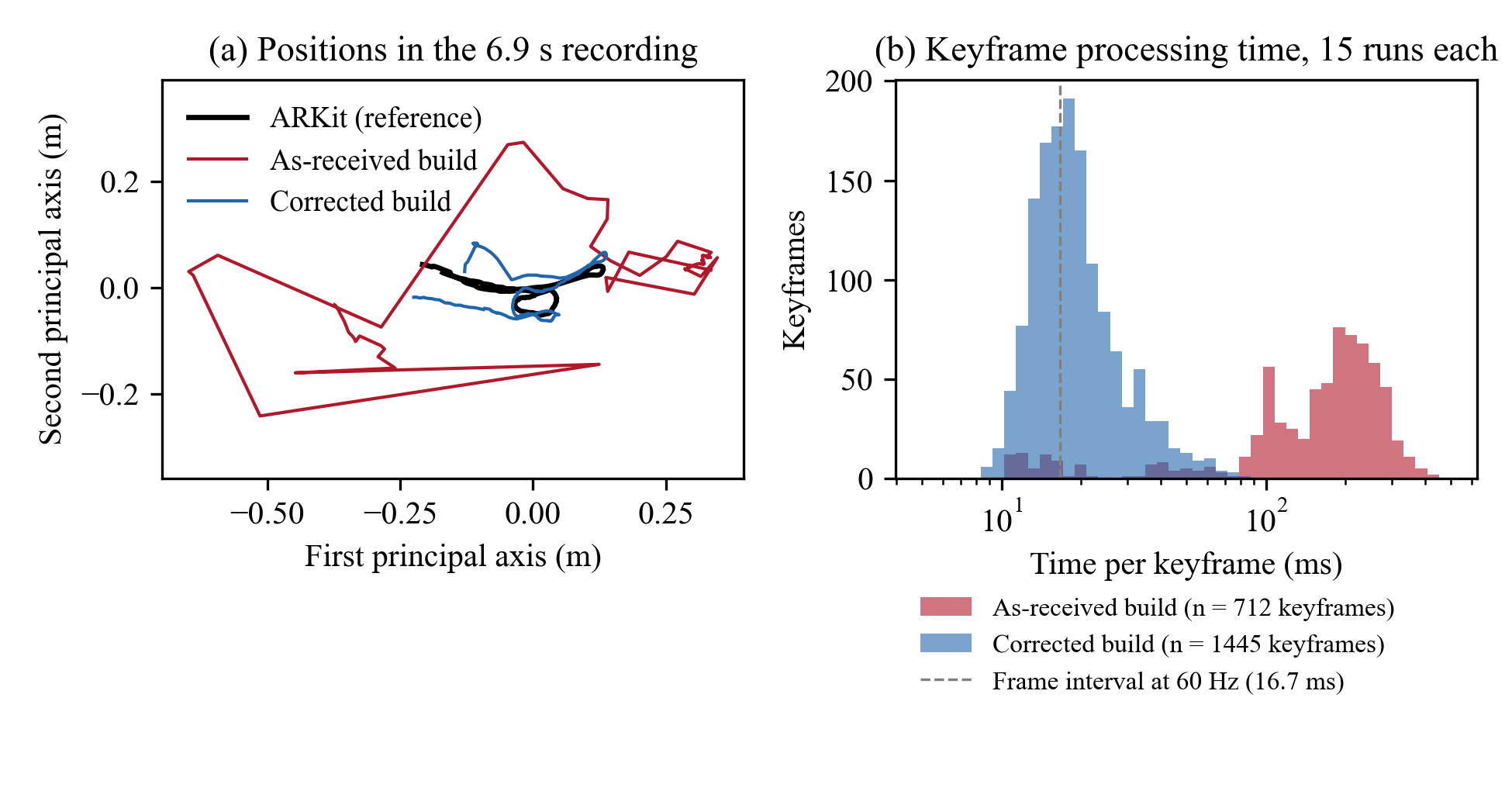}
\caption{On-device replay. (a) Positions for the 6.9 s recording in the plane of the two principal axes of the ARKit path, after rigid alignment (black: ARKit; red: as-received build; blue: corrected build). (b) Time per keyframe, pooled over all keyframes of the 15 runs of each build; the dashed line is the 16.7 ms frame interval of the two 60 Hz recordings. Only keyframes are timed here. Table 4 gives medians of per-run values, which differ slightly from pooled medians.}
\end{figure}

\medskip\noindent\textbf{Which change did what.} Replaying the same recordings on a desktop with every frame processed separates the changes (Table 5). With all frames processed, even the as-received source stayed within 1.6--7.8 cm of ARKit, so most of its error on the phone came from dropped frames. Its path, however, was 3.7--7.8 times as long as the ARKit path: the pose oscillated from keyframe to keyframe. The axis mapping is a correctness fix, and its effect in this replay was mixed: the position difference fell in two recordings and was essentially unchanged in one (3.9 and 4.1 cm), and the path ratio did not improve consistently. The prior term removed the oscillation (path ratio 1.2--1.3) and cut the median iterations in two of the three recordings. Candidate selection with the stopping rule then cut registration time by a factor of 19 to 30 with no change in position difference.

\begin{table}[htbp]
\centering
\caption{Desktop replay of the three phone recordings with every frame processed. Each cell lists the 12.7 s, 6.9 s, and 8.4 s recordings in that order; rows add changes cumulatively.}
\small
\begin{tabular}{|>{\centering\arraybackslash}p{1.12in}|>{\centering\arraybackslash}p{1.12in}|>{\centering\arraybackslash}p{1.12in}|>{\centering\arraybackslash}p{1.12in}|>{\centering\arraybackslash}p{1.12in}|}
\hline
\textbf{Setting} & \textbf{Difference from ARKit, RMS (cm)} & \textbf{Estimated path / ARKit path} & \textbf{Iterations (median)} & \textbf{Registration time (ms)} \\ \hline
As-received source & 1.6, 3.9, 7.8 & 7.8, 3.7, 5.2 & 20, 20, 20 & 100, 144, 176 \\ \hline
+ axis mapping & 1.1, 4.1, 2.3 & 4.9, 4.1, 5.2 & 20, 20, 20 & 84, 143, 169 \\ \hline
+ prior term & 0.8, 3.2, 1.8 & 1.2, 1.2, 1.3 & 11, 8, 20 & 46, 68, 95 \\ \hline
+ candidates and stopping rule & 0.8, 3.2, 1.8 & 1.2, 1.2, 1.2 & 2, 2, 2 & 2.4, 3.1, 3.2 \\ \hline
\end{tabular}
\end{table}

\medskip\noindent\textbf{What this does not show.} The recordings are short, cover one room and one device, and ARKit is the only reference. They show that the corrected build kept up with the sensor in replay on a phone and stayed consistent with ARKit over a few seconds; they do not show accuracy over a room-scale scan, and the corrected build has not been evaluated on the held-out scenes.

\section{Comparison With Open Estimators and Phone Scans}

\noindent This section reports the comparison with open LiDAR--inertial estimators on seven scenes, the effect of the input condition, desktop processing cost, the default ARKit tracker, and reconstructions exported by the phone application.

\medskip\noindent\textbf{Six open estimators had ATE of at least 4.46 m on 41 of 42 runs; Mobile-LIO stayed bounded.} Table 6 gives the ATE of six open LiDAR--inertial estimators and of depth--inertial Mobile-LIO on the three held-out and four development scenes, all given the same depth points and inertial samples and scored against the dataset poses. Mobile-LIO had the lowest ATE on every scene, 0.200--1.973 m. The best other result was 0.806 m (iG-LIO on S1); the remaining 41 scores were at least 4.46 m. The comparison is not controlled: Mobile-LIO clips speed at 0.5 m/s and the others do not, no estimator was retuned for phone depth, iG-LIO and LIO-Livox also showed large errors on some noise-free adapter tests, and DLIO stopped producing output before the end of two recordings. The table therefore shows that estimators configured for dedicated scanners do not transfer to this input as configured; it does not rank the algorithms. Figure 7 shows the position error over time on the held-out scenes after one rigid alignment per estimator. A single alignment spreads late divergence over the whole run, so the curves of the diverged estimators do not show when each one failed; Mobile-LIO stays within a few meters throughout.

\begin{table}[htbp]
\centering
\caption{ATE (m) against the dataset poses for six open LiDAR--inertial estimators and depth--inertial Mobile-LIO given the same phone depth and inertial input. S1--S3 are the held-out scenes; the others are development scenes. DLIO output covered 82\% and 83\% of S2 and S3.}
\small
\begin{tabular}{|>{\centering\arraybackslash}p{0.65in}|>{\centering\arraybackslash}p{0.65in}|>{\centering\arraybackslash}p{0.65in}|>{\centering\arraybackslash}p{0.65in}|>{\centering\arraybackslash}p{0.65in}|>{\centering\arraybackslash}p{0.65in}|>{\centering\arraybackslash}p{0.65in}|>{\centering\arraybackslash}p{0.65in}|}
\hline
\textbf{Scene} & \textbf{DLIO} & \textbf{FAST-LIO2} & \textbf{Faster-LIO} & \textbf{Point-LIO} & \textbf{iG-LIO} & \textbf{LIO-Livox} & \textbf{Mobile-LIO} \\ \hline
S1 & 155.8 & 1,151.7 & 158.0 & 9.09 & 0.806 & 11,521.5 & 0.409 \\ \hline
S2 & 8,069.7 & 4,642.1 & 833.2 & 7.77 & 944.2 & 43,628.5 & 0.200 \\ \hline
S3 & 6,810.7 & 82,003.7 & 3,068.4 & 100.5 & 6,180.8 & 52,938.2 & 1.973 \\ \hline
09c1414f1b & 2,879.2 & 4,907.6 & 30.6 & 5.37 & 5,748.1 & 96,550.9 & 1.105 \\ \hline
0d2ee665be & 12.6 & 129.7 & 208.9 & 4.46 & 375.1 & 10,998.6 & 0.871 \\ \hline
1ada7a0617 & 8,988.0 & 719.7 & 1,582.7 & 6.14 & 244.7 & 19,797.2 & 0.578 \\ \hline
25f3b7a318 & 5.10 & 3,432.1 & 10,880.9 & 5.79 & 3,534.8 & 42,545.1 & 0.526 \\ \hline
\end{tabular}
\end{table}

\begin{figure}[htbp]
\centering
\includegraphics[width=5.60in]{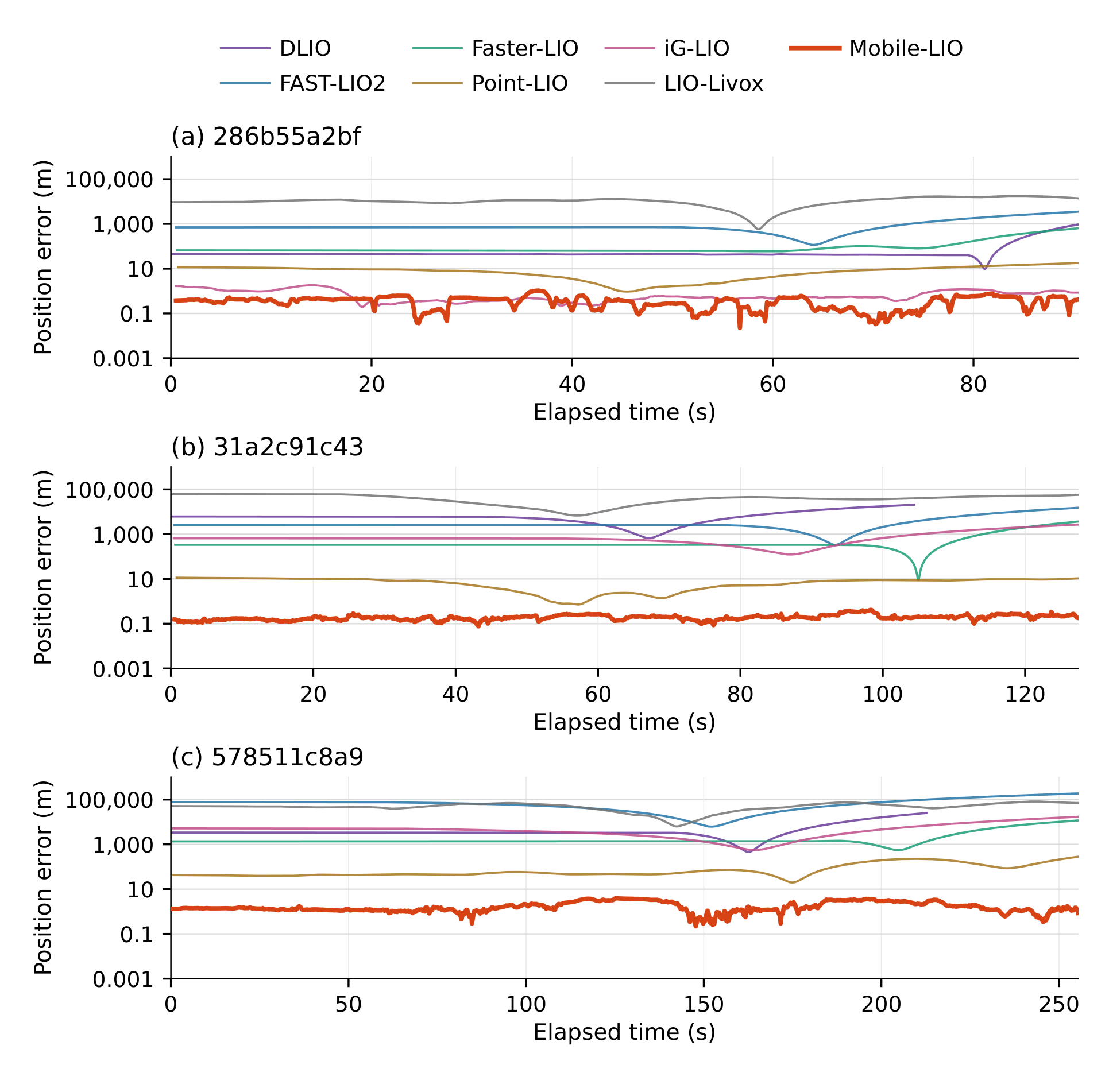}
\caption{Position error over time against the dataset poses on the held-out scenes (S1 to S3 from top), after one rigid alignment per estimator, on a common logarithmic scale. Panel labels 286b55a2bf, 31a2c91c43, and 578511c8a9 are S1, S2, and S3; the DLIO curves end early because its output stopped.}
\end{figure}

\medskip\noindent\textbf{With a 3 m limit, the visual update improved four of seven scenes by more than 0.1 m.} Table 7 gives Mobile-LIO ATE under three input conditions. Comparing the two 3 m columns, which share depth, inertial input, and initialization, the visual update lowered ATE by more than 0.1 m on four scenes (S3 and three development scenes), changed it by less than 0.1 m on S1 and S2, and raised it on one development scene (0.943 to 1.656 m). The 10 m condition also differs from the 3 m conditions in its input frame convention and initialization, so the first two columns do not isolate range.

\begin{table}[htbp]
\centering
\caption{Mobile-LIO ATE (m) against the dataset poses by input condition.}
\small
\begin{tabular}{|>{\centering\arraybackslash}p{1.44in}|>{\centering\arraybackslash}p{1.44in}|>{\centering\arraybackslash}p{1.44in}|>{\centering\arraybackslash}p{1.44in}|}
\hline
\textbf{Scene} & \textbf{Depth and IMU, 10 m limit} & \textbf{Depth and IMU, 3 m limit} & \textbf{Image, depth and IMU, 3 m limit} \\ \hline
S1 & 0.409 & 0.441 & 0.435 \\ \hline
S2 & 0.200 & 0.230 & 0.170 \\ \hline
S3 & 1.973 & 2.920 & 0.590 \\ \hline
09c1414f1b & 1.105 & 0.413 & 0.190 \\ \hline
0d2ee665be & 0.871 & 1.185 & 0.207 \\ \hline
1ada7a0617 & 0.578 & 0.943 & 1.656 \\ \hline
25f3b7a318 & 0.526 & 0.704 & 0.245 \\ \hline
\end{tabular}
\end{table}

\medskip\noindent\textbf{Cost on a desktop.} Table 8 gives the processing cost per frame of the desktop replay on a lightly loaded host, with the error of the same run. On the phone recordings with a 3 m limit, the depth--inertial replay took 188 and 272 ms per frame on S1 and S2 and the replay with the visual update 314 and 366 ms, all slower than the 10 Hz input. Fewer iterations cut the time but raised the error (S1: 0.435 m at 20 iterations, 0.734 m at 5; S2: 0.170 and 0.338 m). On a MID-360 recording, 3 iterations took 69 ms per scan with no loss against 20 iterations (0.152 and 0.162 m). Peak memory was 204--493 MiB. These are whole-process times of the offline replay, including diagnostics and export, and are not the on-device timings of the previous section.

\begin{table}[htbp]
\centering
\caption{Desktop replay cost per saved frame, peak memory, and error for Mobile-LIO. Phone recordings use a 3 m limit and ATE against the dataset poses; the MID-360 recording (KI run\_2) uses the difference from the supplied comparison trajectory and up to 3,200 points per scan.}
\small
\begin{tabular}{|>{\centering\arraybackslash}p{0.76in}|>{\centering\arraybackslash}p{0.76in}|>{\centering\arraybackslash}p{0.76in}|>{\centering\arraybackslash}p{0.76in}|>{\centering\arraybackslash}p{0.76in}|>{\centering\arraybackslash}p{0.76in}|>{\centering\arraybackslash}p{0.76in}|}
\hline
\textbf{Recording} & \textbf{Configuration} & \textbf{Iterations} & \textbf{CPU (ms)} & \textbf{Elapsed (ms)} & \textbf{Error (m)} & \textbf{Peak memory (MiB)} \\ \hline
S1 & Depth and IMU & 20 & 191.5 & 188.1 & 0.441 & 204 \\ \hline
S1 & Image, depth and IMU & 20 & 310.0 & 313.7 & 0.435 & 215 \\ \hline
S1 & Image, depth and IMU & 10 & 215.5 & 213.6 & 0.588 & 212 \\ \hline
S1 & Image, depth and IMU & 5 & 171.7 & 169.4 & 0.734 & 210 \\ \hline
S2 & Depth and IMU & 20 & 277.4 & 271.8 & 0.230 & 266 \\ \hline
S2 & Image, depth and IMU & 20 & 365.0 & 366.4 & 0.170 & 317 \\ \hline
S2 & Image, depth and IMU & 10 & 223.6 & 226.6 & 0.171 & 306 \\ \hline
S2 & Image, depth and IMU & 5 & 154.1 & 152.7 & 0.338 & 294 \\ \hline
MID-360 & 300 residuals & 20 & 408.2 & 399.2 & 0.195 & 464 \\ \hline
MID-360 & 1,500 residuals & 20 & 389.8 & 381.3 & 0.162 & 493 \\ \hline
MID-360 & 1,500 residuals & 5 & 104.1 & 102.7 & 0.167 & 493 \\ \hline
MID-360 & 1,500 residuals & 3 & 69.5 & 68.9 & 0.152 & 492 \\ \hline
\end{tabular}
\end{table}

\medskip\noindent\textbf{ARKit, the default tracker.} ARKit is the tracker that phone scanning applications use by default, so it is the baseline a practitioner already has, without the scale correction applied by the dataset. Table 9 compares its raw poses with the dataset poses on the ten scenes. Because the dataset poses are those same ARKit poses after a scale and a transform, this is not an accuracy measurement against an independent reference; it shows the size of that adjustment. After one rigid alignment the raw poses differed from the dataset poses by 3--17 cm (root mean square) over the ten scenes (3--8 cm on the three held-out scenes), and the raw paths were 1--6\% shorter. The mesh-based evidence on ARKit poses is the surface result: maps fused with the dataset poses lay a median 1.8--2.3 cm from the laser mesh on the held-out scenes, against 8--17 cm with Mobile-LIO poses (Table 3). Those maps show what ARKit poses give with this depth once the dataset's scale correction and mesh alignment are applied. Mobile-LIO and the open estimators received only a rigid alignment, no surface comparison was made for the open estimators, and the accuracy of uncorrected ARKit tracking is not measured here.

\begin{table}[htbp]
\centering
\caption{Raw ARKit poses compared with the dataset poses (the same poses after the dataset's scale and transform) at 10 Hz, after one rigid alignment without scale. Path ratio is the raw ARKit path length divided by the dataset-pose path length. S1--S3 are the held-out scenes.}
\small
\begin{tabular}{|>{\centering\arraybackslash}p{0.76in}|>{\centering\arraybackslash}p{0.76in}|>{\centering\arraybackslash}p{0.76in}|>{\centering\arraybackslash}p{0.76in}|>{\centering\arraybackslash}p{0.76in}|>{\centering\arraybackslash}p{0.76in}|>{\centering\arraybackslash}p{0.76in}|}
\hline
\textbf{Scene} & \textbf{Frames} & \textbf{Duration (s)} & \textbf{RMS difference (cm)} & \textbf{Median difference (cm)} & \textbf{Maximum difference (cm)} & \textbf{Path ratio} \\ \hline
S1 & 911 & 91 & 3.0 & 2.7 & 5.3 & 0.987 \\ \hline
S2 & 1,275 & 127 & 3.3 & 2.5 & 13.6 & 0.945 \\ \hline
S3 & 2,554 & 255 & 8.2 & 7.4 & 18.2 & 0.992 \\ \hline
09c1414f1b & 1,816 & 182 & 6.1 & 5.4 & 12.6 & 0.988 \\ \hline
0d2ee665be & 900 & 90 & 10.4 & 6.6 & 21.7 & 0.994 \\ \hline
13c3e046d7 & 1,536 & 154 & 5.7 & 5.3 & 9.7 & 0.968 \\ \hline
1ada7a0617 & 919 & 92 & 10.1 & 9.0 & 20.0 & 0.981 \\ \hline
21d970d8de & 1,524 & 152 & 12.8 & 11.7 & 23.4 & 0.982 \\ \hline
25f3b7a318 & 1,757 & 176 & 9.7 & 9.2 & 19.6 & 0.939 \\ \hline
27dd4da69e & 4,337 & 434 & 17.3 & 15.5 & 32.6 & 0.977 \\ \hline
\end{tabular}
\end{table}

\medskip\noindent\textbf{Reconstructions.} Figure 8 fuses the same recorded depth over an initial interval of each held-out scene with dataset poses, FAST-LIO2 poses, and Mobile-LIO poses. With FAST-LIO2 poses the room in S2 is visibly distorted, with displaced walls; with Mobile-LIO poses the walls remain separate surfaces, with visible misalignment in S1.

\begin{figure}[htbp]
\centering
\includegraphics[width=6.00in]{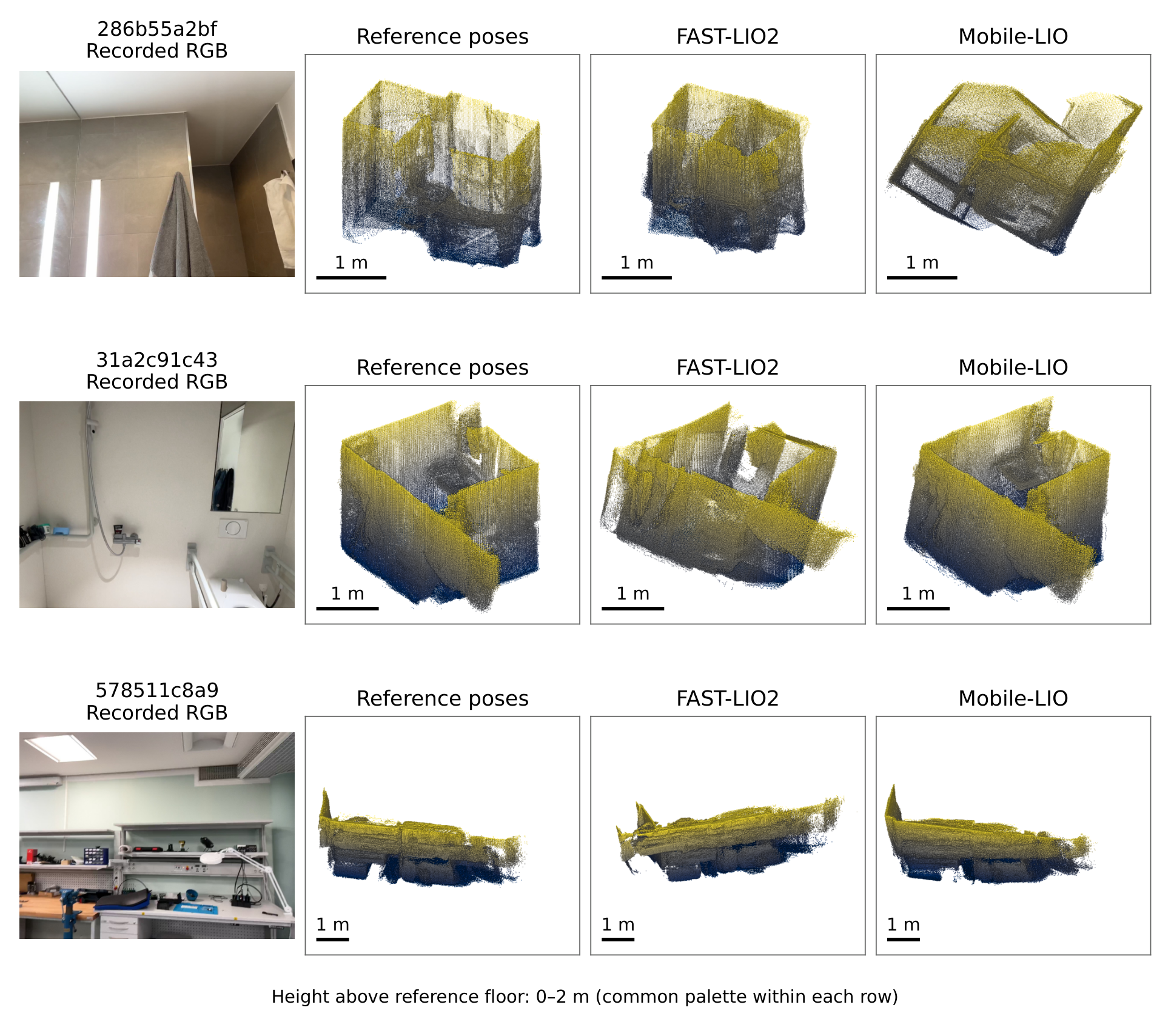}
\caption{Recorded image and recorded depth fused over an initial interval of each held-out scene (S1 to S3 from top; 35.6, 71.8, and 21.8 s) with dataset poses (labeled reference poses), FAST-LIO2 poses, and Mobile-LIO poses. Scene labels 286b55a2bf, 31a2c91c43, and 578511c8a9 are S1, S2, and S3. Color is height above the floor, from 0 m (dark blue) to 2 m (yellow), with a common palette within each row.}
\end{figure}

\medskip\noindent\textbf{Scans exported by the phone application.} Figure 9 shows point clouds exported by the phone application (DepthViz) for a corridor loop, a staircase, a room, and a scene with mirrors, captured on 15 October 2026. They show that the application reconstructs these spaces on the device. They are qualitative: the build installed on that day was not recorded, and no reference was captured for the staircase or the rooms. For the corridor, a Livox MID-360 scanner on a cart recorded the same route (Figure 10); after one rigid alignment the median distance from the phone export to the scanner map was 0.77 m over the whole extent and 0.26 m within the wall band, so the loop shape is recovered but not at survey accuracy. The scanner map is itself an odometry result without loop closure.

\begin{figure}[htbp]
\centering
\includegraphics[width=6.40in]{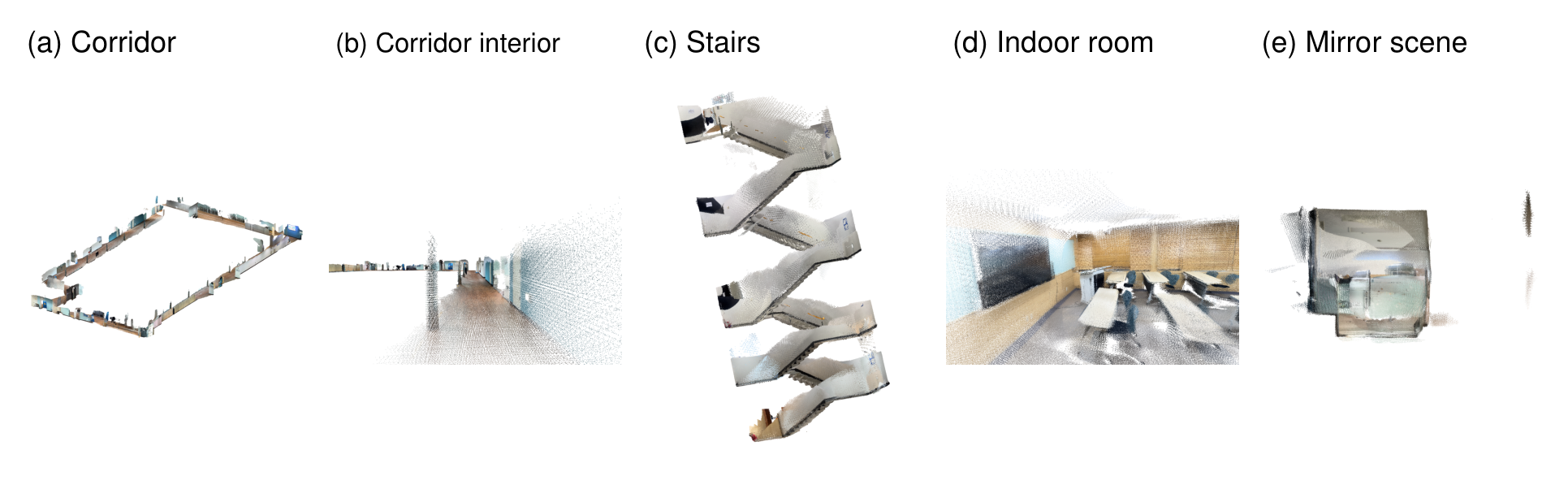}
\caption{Point clouds exported by the phone application: (a) corridor loop, (b) view along the corridor, (c) staircase, (d) room, and (e) scene with mirrors. Original colors; views are not to a common scale.}
\end{figure}

\begin{figure}[htbp]
\centering
\includegraphics[width=6.00in]{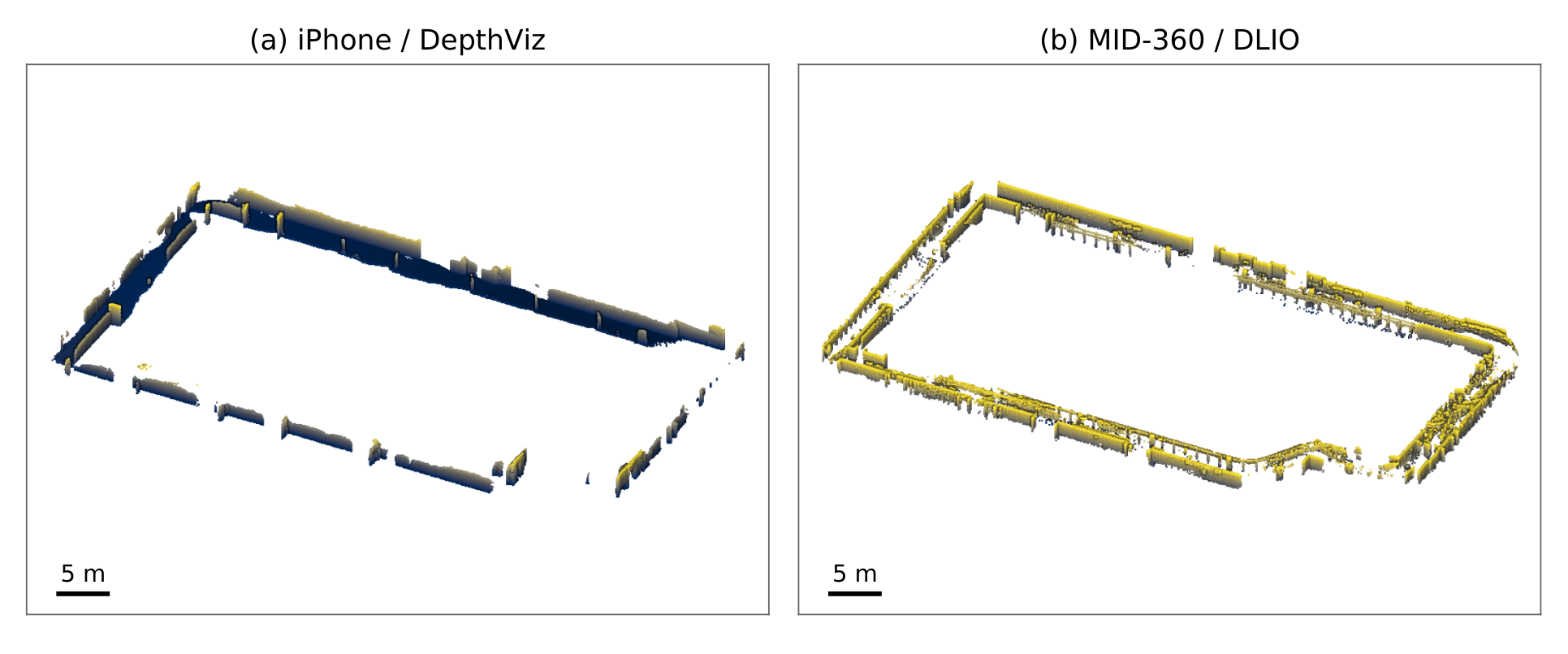}
\caption{The same corridor from (a) the phone export (DepthViz is the phone application) and (b) a Livox MID-360 scanner fused with DLIO odometry, after one rigid alignment. Color is height, from 0.3 m below (dark blue) to 2.2 m above (yellow) a fixed reference floor, on a palette common to both panels, with a common camera and scale; points within 1.5 m of the route are shown. Heights are not flattened along the path, so the difference in overall color between the panels indicates a height offset that remains between the two maps after the alignment.}
\end{figure}

\section{Discussion and Limitations}

\noindent For practitioners the budget gives three points. First, range. Recorded depth had a median error of 1--5 cm within 3 m and 20--54 cm at 5--7 m, and about 90\% of pixels in these recordings lay within 3 m. S3 was the exception, with 16.6\% of its pixels beyond 4 m. The depth data suggest scanning surfaces from within about 3 m, but the effect of a range limit on pose was mixed and this was not tested as a protocol. Second, tracking. With ARKit poses aligned to the mesh the maps lay a median 2 cm from it, and Mobile-LIO poses multiplied that by four to eight, so tracking, not depth, separates the two. The aligned poses include a scale correction that a field user does not have, so a scan should be checked against a few taped dimensions or control points before it is used. Third, fitness for purpose. The ARKit-pose maps indicatively correspond to LOA10 (outlier filtering and error in the mesh could change the 17--27\% share beyond 5 cm), which may serve space inventory, visual context, and coarse progress records but not dimensioned as-built modeling or clearance checks. The Mobile-LIO maps, with a median of 8--17 cm and a 90th percentile of 25--64 cm, are not suitable for any dimensioned use.

For research, the controlled experiment shows that the estimator reaches centimeter-level deviation from the reference on clean depth in the one scene tested, but not which depth defect to remove first. No independent ground-truth trajectory was available, so the trajectory accuracy of ARKit itself is not measured here; its maps are, through the mesh.

Inspectability paid off in diagnosis. The logs showed that the initial gyroscope-bias state (0.019--0.136 rad/s in nine scenes) far exceeded the bias measured against dataset poses (below 0.004 rad/s) because no recording starts at rest, and they located the onset of error growth in S3.

Several limits apply. The on-device check covers one room, one phone, and recordings of 7--13 s, with ARKit as the only reference. The exported scans are qualitative, and the corridor comparison uses an odometry result, which drifts, as its reference. Trajectory metrics use an ARKit-derived reference. The depth comparison inherits any error in the dataset poses and mesh. The evaluation covers three held-out scenes; the phone configuration and the controlled experiment were run after the first held-out results and are post hoc. The speed cap clipped 13--49\% of propagation steps on the seven development scenes. Nearest-surface distance does not penalize duplicated structure near another mesh surface, and the maps were placed by a best-fit trajectory alignment that is unavailable in field use. Unlike common practice in this literature, the maps were not registered to the reference by ICP and room dimensions were not measured. Replay used constant confidence, so confidence rejection on native data was not tested. The estimator has no loop closure and was evaluated only on single indoor scenes.

\section{Conclusion}

\noindent ARKit gives applications depth and poses but not the means to inspect how they were produced. This paper examined whether room-scale scanning is feasible with an open depth--inertial pipeline that uses no ARKit poses, and measured its distance from the closed pipeline against a laser-scanned reference. With ARKit poses that the dataset scales and aligns to the mesh, a step a field user cannot take, recorded depth produced maps about 2 cm from the laser mesh at the median, though 17--27\% of points lay beyond 5 cm. Recorded iPhone depth had a median error of a few centimeters within 3 m, with a tail, and was unreliable beyond 5 m. By the 95\% criterion the ARKit-pose maps indicatively correspond to the coarsest USIBD accuracy level, while Mobile-LIO poses raised the median distance to 8--17 cm. Recorded depth was most accurate within about 3 m, although whether restricting range improves a scan was not tested, and a scan should be verified against known dimensions before use. Mobile-LIO deviated from the ARKit-derived reference by 0.2--2.0 m, and noise-free depth brought it to centimeter-level deviation in the scene where its error was largest. Six open LiDAR--inertial estimators, run as configured for dedicated scanners and not retuned, had ATE of at least 4.46 m in 41 of 42 runs on this input (an uncontrolled comparison; Mobile-LIO has a speed cap they lack), and qualitative phone exports from an unrecorded build showed a corridor, a staircase, and rooms reconstructed on the device (corridor median 0.77 m from a scanner map). A build with corrected phone code kept up with the sensor in 14 of 15 replay runs without display load on one phone (18 ms per keyframe; 83\% and 95\% of frames in two live scans), but its room-scale accuracy is untested and the dataset results use the estimator before those corrections. Future work will identify which depth defects matter, evaluate raw ARKit poses without the dataset's scale correction, measure room dimensions, and extend the held-out set.

\section*{Appendix: Estimator Settings}

\noindent Table 10 lists the settings of the estimator used for every result in the Results section.

\begin{table}[htbp]
\centering
\caption{Settings of the Mobile-LIO estimator evaluated on the dataset.}
\small
\begin{tabular}{|>{\centering\arraybackslash}p{3.01in}|>{\centering\arraybackslash}p{3.01in}|}
\hline
\textbf{Item} & \textbf{Value} \\ \hline
Depth sampling & every 4th pixel of the 256 $\times$ 192 map (at most 3,072 points) \\ \hline
Input rejection & confidence level 0; nonfinite; range above 10 m (depth--inertial) or 3 m (phone configuration) \\ \hline
Point reduction & one centroid per 3 cm voxel \\ \hline
State & orientation, position, velocity, gyroscope bias, accelerometer bias, gravity (18 error states) \\ \hline
Initialization & 30 motion-sensor samples; gravity magnitude set to the measured mean; initial variances 10\textsuperscript{$-$4}, gravity 1.0 \\ \hline
Process noise densities & gyroscope 0.01 and accelerometer 0.02 (rad/s and m/s\textsuperscript{2} per root hertz); bias random walks 0.001 and 0.005 in the corresponding units \\ \hline
Speed cap & 0.5 m/s on the propagated velocity \\ \hline
Estimation map & hash of 10 cm cells, up to 20 points per cell \\ \hline
Plane fit & points within a 20 cm radius, at least 6, smallest-to-middle eigenvalue ratio at most 0.3 \\ \hline
Residual gate and noise & reject above 10 cm; standard deviation 2 cm; weight 1 or 0.5 by confidence \\ \hline
Residual budget & at most 100 per update, evenly sampled \\ \hline
Repeated correction & up to 20 iterations; stop below 10\textsuperscript{$-$6}; no prior term; covariance updated once after the last iteration \\ \hline
Update rejection & total correction above 1 m or 0.5 rad restores the entry state \\ \hline
Keyframes and map & keyframe at 5 cm or 2$^\circ$; 5 seeding frames; map reset after 10 consecutive failed updates \\ \hline
Visual update (phone configuration) & up to 200 Shi--Tomasi corners tracked by pyramidal Lucas--Kanade flow with a 1.5 pixel forward--backward check; reprojection residuals with 2 pixel standard deviation; needs at least 5 tracked features \\ \hline
Extrinsics & identity between camera and motion sensor \\ \hline
\end{tabular}
\end{table}

\section*{References}
\begin{list}{}{\setlength{\leftmargin}{0.5in}\setlength{\itemindent}{-0.5in}\setlength{\itemsep}{0pt}\setlength{\parsep}{0pt}}
\item Abdelghany, A. A., D. D. Lichti, P. Dawson, and S. Jabari. 2026. ``Evaluating iPhone-based 3D scanning applications for heritage documentation: Controlled experiments and future directions.'' \textit{Int. Arch. Photogramm. Remote Sens. Spatial Inf. Sci.} XLIX-M-1-2026: 1--10.
\item Abdel-Majeed, H. M., I. F. Shaker, A. M. Abdel-Wahab, and A. A. D. I. Awad. 2024. ``Indoor mapping accuracy comparison between the Apple devices' LiDAR sensor and terrestrial laser scanner.'' \textit{HBRC J.} 20 (1): 915--931.
\item Askar, C., and H. Sternberg. 2023. ``Use of smartphone lidar technology for low-cost 3D building documentation with iPhone 13 Pro: A comparative analysis of mobile scanning applications.'' \textit{Geomatics} 3 (4): 563--579.
\item Bai, C., T. Xiao, Y. Chen, H. Wang, F. Zhang, and X. Gao. 2022. ``Faster-LIO: Lightweight tightly coupled lidar-inertial odometry using parallel sparse incremental voxels.'' \textit{IEEE Rob. Autom. Lett.} 7 (2): 4861--4868.
\item Chen, K., R. Nemiroff, and B. T. Lopez. 2023. ``Direct LiDAR-inertial odometry: Lightweight LIO with continuous-time motion correction.'' In \textit{Proc., IEEE Int. Conf. on Robotics and Automation}, 3983--3989. New York: IEEE.
\item Davletshina, D., V. K. Reja, and I. Brilakis. 2026. ``Constructing and maintaining road geometric digital twins.'' \textit{J. Comput. Civ. Eng.} 40 (1): 03125002.
\item D\'{\i}az-Vilari\~no, L., H. Tran, E. Fr\'{\i}as, J. Balado, and K. Khoshelham. 2022. ``3D mapping of indoor and outdoor environments using Apple smart devices.'' \textit{Int. Arch. Photogramm. Remote Sens. Spatial Inf. Sci.} XLIII-B4-2022: 303--308.
\item Ersoz, A. B., and F. Bosch\'e. 2026. ``A comparative study of mobile device and terrestrial laser scanning-based scan-to-BIM workflows.'' In \textit{Proc., 43rd Int. Symp. on Automation and Robotics in Construction}, 2242--2249. IAARC.
\item He, D., W. Xu, N. Chen, F. Kong, C. Yuan, and F. Zhang. 2023. ``Point-LIO: Robust high-bandwidth light detection and ranging inertial odometry.'' \textit{Adv. Intell. Syst.} 5 (7): 2200459.
\item Jang, M., and H.-C. Yu. 2026. ``Mobile-LIO: Lightweight LiDAR-inertial odometry with real-time correction for commodity smartphone LiDAR.'' In \textit{SIGGRAPH Asia 2026 Posters}, 1--3. New York: ACM.
\item Jin, L., and C. Ye. 2023. ``Visual-LiDAR-inertial odometry: A new visual-inertial SLAM method based on an iPhone 12 Pro.'' In \textit{Proc., IEEE/RSJ Int. Conf. on Intelligent Robots and Systems}, 1511--1516. New York: IEEE.
\item Keitaanniemi, A., P. R\"onnholm, A. Kukko, and M. T. Vaaja. 2023. ``Drift analysis and sectional post-processing of indoor simultaneous localization and mapping (SLAM)-based laser scanning data.'' \textit{Autom. Constr.} 147 (March): 104700.
\item Luetzenburg, G., A. Kroon, and A. A. Bj{\o}rk. 2021. ``Evaluation of the Apple iPhone 12 Pro LiDAR for an application in geosciences.'' \textit{Sci. Rep.} 11 (1): 22221.
\item Osadcha, I., A. Jurelionis, and P. Fokaides. 2023. ``Geometric parameter updating in digital twin of built assets: A systematic literature review.'' \textit{J. Build. Eng.} 73 (August): 106704.
\item Teo, T.-A., and C.-C. Yang. 2023. ``Evaluating the accuracy and quality of an iPad Pro's built-in lidar for 3D indoor mapping.'' \textit{Dev. Built Environ.} 14 (April): 100169.
\item Teppati Los\`e, L., A. Spreafico, F. Chiabrando, and F. Giulio Tonolo. 2022. ``Apple LiDAR sensor for 3D surveying: Tests and results in the cultural heritage domain.'' \textit{Remote Sens.} 14 (17): 4157.
\item Treccani, D., A. Adami, and L. Fregonese. 2024. ``Assessing the effectiveness of LiDAR-based apps on Apple devices to survey indoor and outdoor medium sized areas.'' \textit{Int. Arch. Photogramm. Remote Sens. Spatial Inf. Sci.} XLVIII-2/W8-2024: 431--438.
\item Trzeciak, M., K. Pluta, Y. Fathy, L. Alcalde, S. Chee, A. Bromley, I. Brilakis, and P. Alliez. 2023. ``ConSLAM: Construction data set for SLAM.'' \textit{J. Comput. Civ. Eng.} 37 (3): 04023009.
\item USIBD (U.S. Institute of Building Documentation). 2016. \textit{Level of accuracy (LOA) specification guide: Document C120, version 2.0}. Tustin, CA: USIBD.
\item Xu, W., Y. Cai, D. He, J. Lin, and F. Zhang. 2022. ``FAST-LIO2: Fast direct LiDAR-inertial odometry.'' \textit{IEEE Trans. Rob.} 38 (4): 2053--2073.
\item Yang, Y., Y.-T. Chen, C. Hancock, N. A. S. Hamm, and Z. Zhang. 2024. ``A novel approach for as-built BIM updating using inertial measurement unit and mobile laser scanner.'' \textit{Remote Sens.} 16 (15): 2743.
\item Yeshwanth, C., Y.-C. Liu, M. Nie{\ss}ner, and A. Dai. 2023. ``ScanNet++: A high-fidelity dataset of 3D indoor scenes.'' In \textit{Proc., IEEE/CVF Int. Conf. on Computer Vision}, 12--22. New York: IEEE.
\end{list}
\end{document}